\documentclass[runningheads]{llncs}
\usepackage{graphicx}
\usepackage{comment}
\usepackage{amsmath,amssymb} 
\usepackage{color}

\begin{document}
\pagestyle{headings}
\mainmatter
\def\ECCVSubNumber{3586}  

\title{AssembleNet++: Assembling Modality Representations via Attention Connections}

\titlerunning{AssembleNet++}
%

\author{Michael S. Ryoo\inst{1,2} \and
AJ Piergiovanni\inst{1} \and
Juhana Kangaspunta\inst{1} \and \\Anelia Angelova\inst{1}}
\authorrunning{M. S. Ryoo et al.}
%
\institute{Robotics at Google \and
Stony Brook University\\
\email{\{mryoo,ajpiergi,juhana,anelia\}@google.com}}
\maketitle

\begin{abstract}
We create a family of powerful video models which are able to: (i) learn interactions between semantic object information and raw appearance and motion features, and (ii) deploy attention in order to better learn the importance of features at each convolutional block of the network. A new network component named \emph{peer-attention} is introduced, which dynamically learns the attention weights using another block or input modality. Even without pre-training, our models outperform the previous work on standard public activity recognition datasets with continuous videos, establishing new state-of-the-art. We also confirm that our findings of having neural connections from the object modality and the use of peer-attention is generally applicable for different existing architectures, improving their performances. We name our model explicitly as AssembleNet++. The code will be available at: \url{https://sites.google.com/corp/view/assemblenet/}

\end{abstract}

\section{Introduction}

Video understanding is a fundamental problem in vision with many novel approaches proposed recently.
While many advanced neural architectures have been used for video understanding~\cite{carreira2017quo,tran2018closer}, including two-stream and multi-stream ones~\cite{carreira2017quo,feichtenhofer2018slowfast,ryoo2019assemblenet}, 
learning of interactions between raw input modalities (e.g., RGB and motion) and semantic input modalities such as objects in the scene (e.g., persons and objects) have been limited.



Inspired by previous work, e.g. AssembleNet architectures for videos~\cite{ryoo2019assemblenet} and RandWire architectures for images~\cite{randwire} which proposed random or targeted connectivity between layers in a neural network, we create a family of powerful video models that explicitly learn interactions between spatial object-specific information and raw appearance and motion features. In particular,  inter-block attention connectivity is searched for to best capture the interplay between different modality representations.


The main technical contributions of this paper include:
\begin{enumerate}
\item Optimizing neural architecture connectivity for object modality fusion. We discover that models with `omnipresent' connectivity from object input allows the best multi-modal fusion.

\item Learning of video models with \emph{peer-attention} on the connections. We newly introduce an one-shot model formulation to efficiently search for architectures with better peer-attention connectivity.
\end{enumerate}

We test the approach extensively on challenging video understanding datasets, showing notable improvements: compared to the baseline backbone architecture we use, our new one-shot attention search model with object modality obtains +12.6\% on Charades classification task and +6.22\% on Toyota Smarthome dataset. Our approach also outperforms reported numbers of existing approaches on both datasets, establishing new state-of-the-art.




\section{Previous work}

\subsubsection{Video CNNs} 

Convolutional neural network (CNNs) for videos~\cite{simonyan2014two,carreira2017quo,Feichtenhofer2016Spatiotemporal,feichtenhofer2018slowfast,qiu2017learning,Wang2016apperance,kay2017kinetics,xie2018rethinking,tran2014c3d,tran2018closer} are a popular approach to video understanding, for example, solutions, such as 3D Video CNNs~\cite{taylor2010conv,3dconv,tran2015learning,carreira2017quo,tran2014c3d,hara2017learning}, (2+1)D CNNs~\cite{tran2018closer} or even novel architecture searched models~\cite{nekrasov2019architecture,piergiovanni2018evolving,ryoo2019assemblenet} are widely used.
Action recognition has also been the topic of intense research~\cite{girdhar2017actionvlad,wang2011action}.

\subsubsection{Action recognition with objects}

Action recognition with objects has been traditionally studied years back~\cite{moore1999object}.
The presence of specific objects in video frames, has been shown to be important for video recognition, even in the context of advanced feature learned by deep neural models, e.g., Sigurdsson et al.~\cite{sigurdsson2017what}; they are useful even if provided as a single label per frame. This is not surprising as many of the activities, e.g. `speaking on the phone', or `reading a book' are primarily determined by the objects themselves. Furthermore, clues about the location of persons, e.g., by 2D human pose has also been shown to be beneficial~\cite{das2019toyota}. 
Recent video CNNs have also tried to integrate object-related information, from segmentation~\cite{baradel2018object,ray2018SOA} or pre-training from image datasets~\cite{diba2019holistic}. One-time late (or intermediate) fusion of object representation with RGB and flow representations has been widely used (e.g., \cite{ma2016first}). Ji et al. \cite{ji2020genome} modeled scene relations on top of video CNNs using graph neural networks, for better usage of object information. However, we are not aware of any prior work that `learns' the connectivity between among input modalities including object information, as we do in this paper.

\subsubsection{Attention}

Use of attention within CNNs have been widely studied. Vaswani et al. \cite{vaswani2017attention} investigated different forms and applications of attention while focusing on self-attention. Hu et al. \cite{hu2018squeeze} introduced Squeeze-and-Excitation, which is a form of channel-wise self-attention. Researchers also developed other forms of channel-wise self-attention \cite{woo2018cbam,fu2019dual,huang2019ccnet,li19em,wang2019eca}, often together with spatial self-attention. Attention was also applied to video CNN models \cite{piergiovanni2017learning,long2018attention,das2019toyota}. However, we are not aware of prior work explicitly searching for inter-block attention connectivity (i.e., peer-attention) as we do in this paper.

\subsubsection{Neural architecture search}

Neural Architecture Search (NAS) is the concept of automatically finding superior architectures based on training data \cite{zoph2017neural,zoph2018nas,liu2018progressive,real2019amoeba,tan2019efficient}. Multiple different strategies including learning of reinforcement learning controller (e.g., \cite{zoph2017neural,zoph2018nas} as well as evolutionary algorithms (e.g., \cite{real2019amoeba}) have been developed for NAS. In particular, one-shot differentiable architecture search \cite{bender2018understanding,liu2019darts} has been successful as it does not require a massive amount of model training. RandWire network \cite{randwire} could also be interpreted as a form of differentiable architecture search, as it learns weights of (random) connections to minimize the classification loss.

However, architecture search for neural attention connectivity has been very limited. Ahmed and Torresani \cite{ahmed2017connectivity} searched for layer connectivity and Ryoo et al. \cite{ryoo2019assemblenet} searched for multi-stream connectivity for video CNNs, but they were without any attention learning which becomes a crucial component when we have a mixture of input modalities. We believe this paper is the first paper to search for models with attention connectivity.

\section{Approach}

\subsection{Preliminaries}

This section describes the video CNN architecture framework, which will be used as a base for developing our approach.

We here adopt a multi-stream, multi-block architecture design from AssembleNet~\cite{ryoo2019assemblenet}. AssembleNet design allows learning of connections between modalities and their intermediate features. This architecture is similar to other two-stream models~\cite{carreira2017quo,feichtenhofer2018slowfast}, but is more flexible in two ways: 1) it allows the use of more than two streams, and 2) it allows connections to be formed (and potentially learned) between individual blocks of the neural architecture.

More specifically, the architecture we use has multiple input blocks, each corresponding to an input modality. The network blocks have a structure inspired by ResNet architectures \cite{resnet}. Each input block is composed of a small number of pooling and convolutional layers attached directly on top of the input. 
The input blocks are then connected to network blocks at the next level. 
We follow the (2+1)D ResNet block structure from \cite{tran2018closer}, where each module is composed of one 1D temporal conv., one 2D spatial conv., and one 1x1 conv. layer. A block is formed by repeating the (2+1)D residual module multiple times. This allows a fair and direct comparison between our approach and previous models using the same module and block \cite{tran2018closer,feichtenhofer2018slowfast,ryoo2019assemblenet}.

Each network block (or block for short) can be connected to any block from any modality at the next level, including its own. 
Blocks are organized at levels so that connections do not form cycles.
Connections can also be formed to skip levels.
We note that since many connections between blocks are formed early, the neural blocks themselves will often contain information from many input modalities as early as the first level of the network.

Figure~\ref{fig:per-connection-ratio} (a) shows one example architecture, where the structure of the network and example connectivity can be seen.

\subsection{Input modalities and semantics}
\label{sec:semantics}

In addition to the standard raw RGB video input, motion information is added as a separate modality. More specifically, optical flow, either pre-computed for the dataset~\cite{zach2007duality}, or trained on the fly~\cite{fan2018e2e,piergiovanni2018representation}, has been shown to be a crucial input for achieving better accuracy across the board~\cite{carreira2017quo}.

We here propose to use object segmentation information as a separate `object' modality. 
Objects and their locations provide semantics information which conveys useful information about activities in a video.
Crucially here, semantic information is incorporated in the full architecture so that it is able to interact with other modalities and the intermediate features from them (as described more in Sections \ref{subsec:connections} to \ref{subsec:oneshot}), to maximize its utilization for the best representation.


\subsubsection{Input block details}
We construct an input block for each input modality. 
Each input block is composed of one pooling and up to two convolutional layers for raw RGB and optical flow inputs, and just one pooling layer for semantic object inputs applied directly on top of the inputs. In the object input block, a segmentation mask having an integer class value per pixel is converted into a HxWx$C_O$ tensor using one-hot operation, where $C_O$ is the number of object classes. The segmentation masks are obtained from a model trained on a non-related image-based  dataset.

\subsection{Learning weighted connections}
\label{subsec:connections}

Blocks in the network can potentially form connections with one or more blocks.
While connectivity and the strength of the connectivity could also be hand-coded, we formulate our networks so that they are learnable.


Let $G$ be the connectivity graph of the network where $(j, i)$ specifies that there is a connection from the $j$th block to $i$th block. We allow each block to receive its inputs from multiple different input blocks as well as intermediate convolutional blocks, and generates an output. Specifically, we formulate the input of the block as a weighted summation over multiple connections where we learn one weight for each connection.
\begin{equation}
x_i^{in} = \sum_{(j,i) \in G} \sigma(w_{ji}) \cdot x_j^{out}
\end{equation}
where $i$ and $j$ are block indexes, $x_i^{in}$ corresponds on the final input to the $i$th block, and $x_i^{out}$ corresponds to the output of the block. $\sigma$ is a sigmoid function.

Learning of the connection weights together with the other convolutional layer parameters with the standard back propagation allows the network to optimize itself on which connections to use and which to not based on the training data. In our approach, this is done by initially connecting every possible blocks in the graph while using the block levels to avoid cycles, and then learning them. We consider every connection $(j, i)$ from the $j$th block to $i$th block as valid as long as $L(j) < L(i)$ where $L(i)$ indicates the level of the block.

\subsection{Attention connectivity and peer-attention}
\label{sec:attention}

In addition to having and learning static weights per connection, we use attention to dynamically control the behavior of each connection. The intuition is that objects and activities are correlated, and using attention allows the model to focus on important objects based on motion context and vice versa. 
For instance, motion features of `drinking' could suggest another network stream to focus more on objects related to such motion (e.g., `cups' and `bottles').


We formulate our connectivity graph $G$ to have one more component for each edge: $((j, i), k)$, where $k$ is the convolutional block influencing the connection $(j, i)$ via attention. 
A channel-wise attention is used to implement this behavior. Let $C_i$ be the size of the input channel of block $i$. For each connection $(j, i)$, the attention vector of size $C_i$ is computed per frame as:
\begin{equation}
A_i(x) = [a_1, \ldots, a_{C_i}] = \sigma(f(\mathrm{GAP}(x)))
\end{equation}
where $f$ is a function (one fully connected layer in our case) mapping a vector to a vector of size $C$. $\mathrm{GAP}$ is the global average pooling over spatial resolution in the input tensor, making $\mathrm{GAP}(x)$ to have a form of a vector per frame.

Using $A_i(x)$, the input for each block $i$ is computed by combining every connection $(j, i)$ while considering its attention from block $k$:
\begin{equation}
x_i^{in} = \sum_{((j,i),k) \in G} \sigma(w_{ji}) \cdot (A_i(x_k^{out}) \cdot x_j^{out}).
\label{eq:attention}
\end{equation}
The simplest special case of our attention is self-attention, which is done by making $x_k$ and $x_j$ to be identical. In this form, the usage of attention becomes similar to Squeeze-and-Excitation \cite{hu2018squeeze}.

Importantly, in our approach, we learn to select different $x_k$ where $x_k \neq x_j$, which we discuss more in the following subsection. Attention with $x_k \neq x_j$ implies that the channels to use for the connection is dynamically decided based on another input modality and peer blocks. We more explicitly name this approach as \emph{peer-attention}. 
In principle, we define a `peer' as any block $p$ that could potentially be connected to $i$.
In our formulation where the convolutional blocks are organized into multiple levels (to avoid cycles), the set of peers $P$ for a connection $(j, i)$ is computed as $P_{(j, i)} = \{p ~|~ L(p) < L(i)\}$ where $L(p)$ indicates the level of the block $p$.
We consider the attention connection $((j, i), k)$ to be valid as long as $k \in P_{(j, i)}$.

\begin{figure}
  \centering
  \includegraphics[width=1.0\columnwidth]{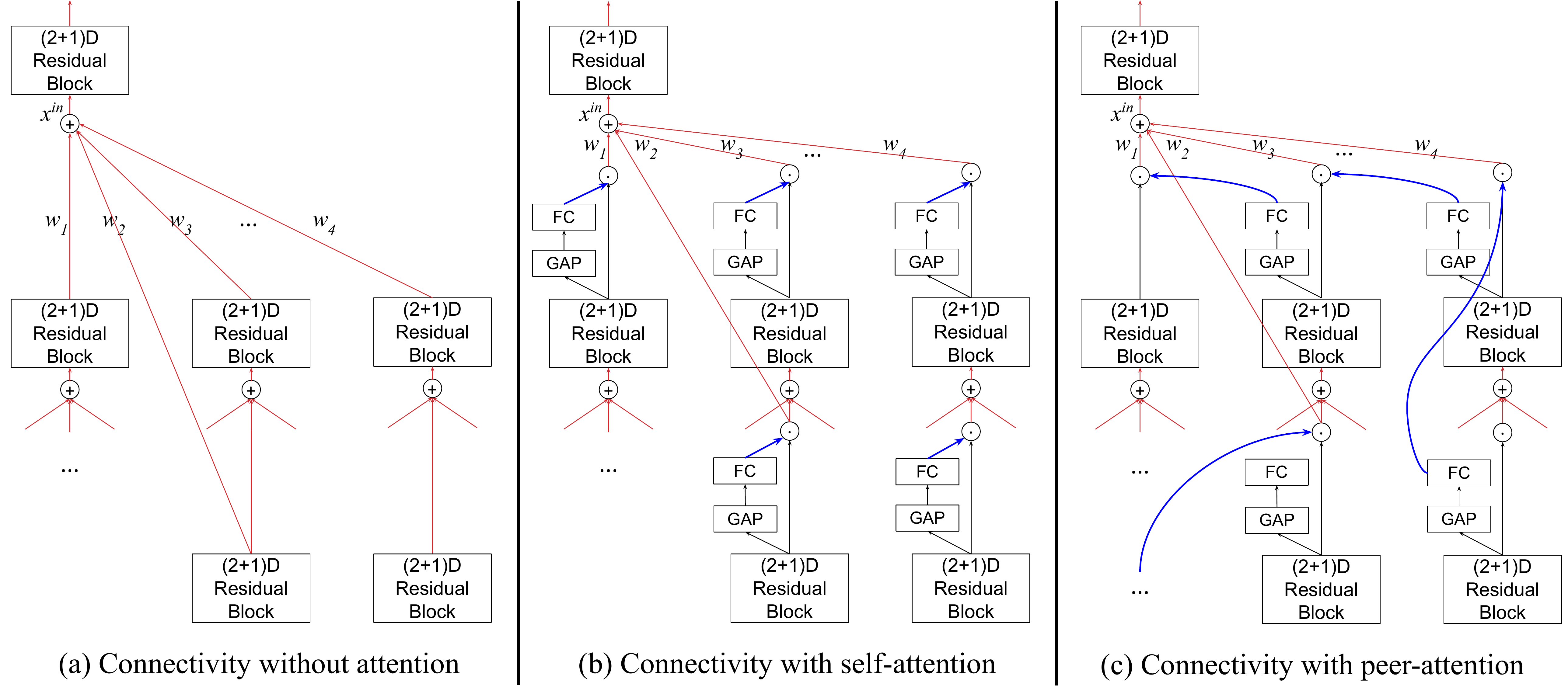}
  \caption{Examples of convolutional block connectivity (a) without attention, (b) with self-attention, and (c) with peer-attention. Red lines indicate weighted connections from Section \ref{subsec:connections}.
  Blue curves specify the attention connectivity. GAP is global average pooling and, FC is a fully connected layer. Our attention is channel-wise attention, and it is applied per frame.}
  \label{fig:peer-attention}
\end{figure}

Figure \ref{fig:peer-attention} compares connectivity without attention and connectivity with self- and peer-attention.

\subsection{One-shot attention search model}
\label{subsec:oneshot}

Given a set of convolutional blocks, instead of hand-designing peer-attention connections, we search for the attention connectivity. Our new one-shot attention search model is introduced, which optimizes the model's peer-attention configuration directly based on training data.




Our one-shot attention search model is formulated by combining attention from all possible peer blocks for each connection with learnable weights.
The idea is to enable the model to soft-select the best peer for each block by learning differentiable weights, maximizing the recognition performance. 
All possible attention connectivity is considered as a consequence, and the searching is done solely based on the standard backpropagation.





\begin{figure}
  \centering
  \includegraphics[width=0.78\columnwidth]{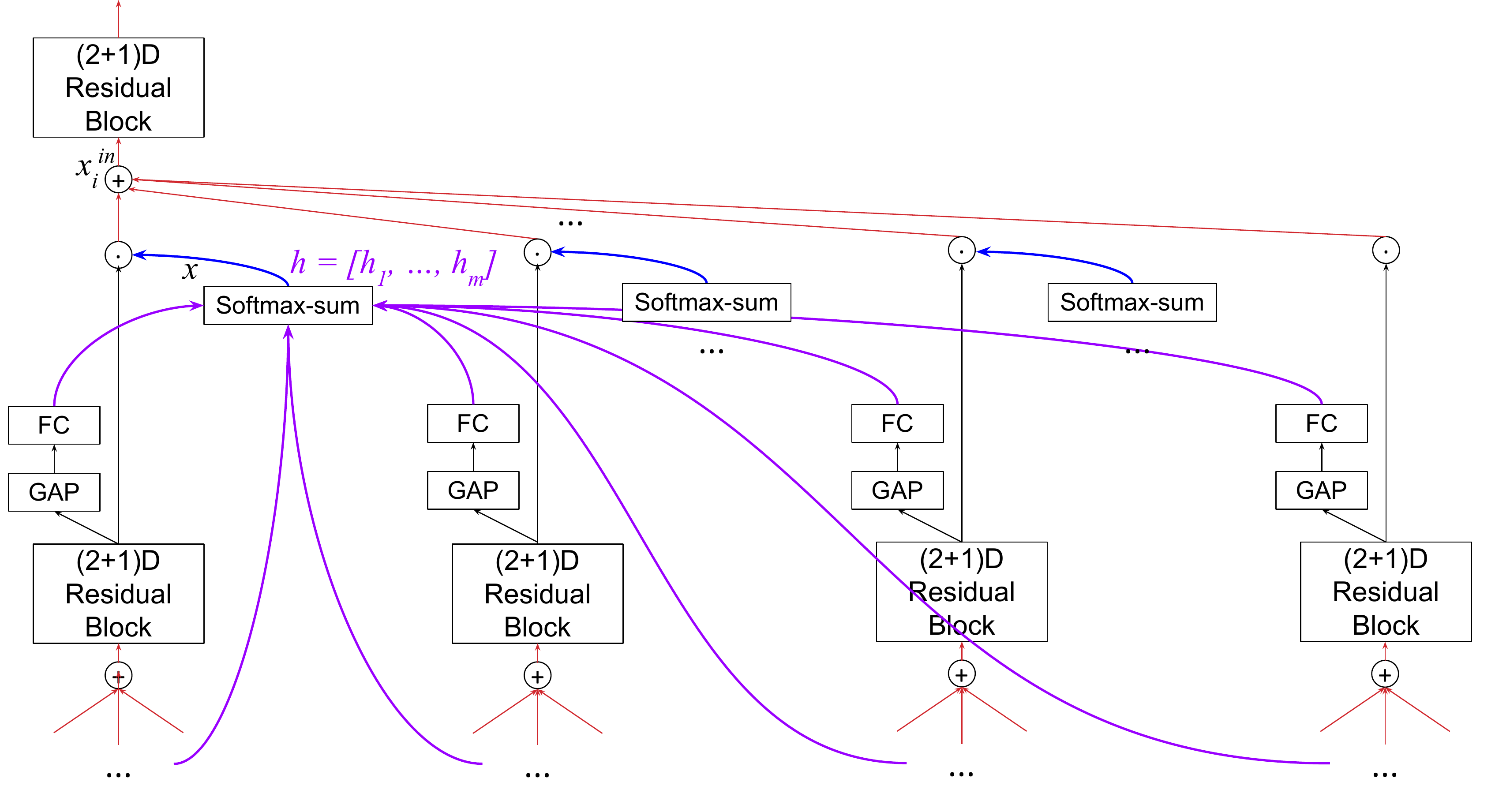}
  \caption{Visualization of our one-shot attention search model. Magenta connections illustrate weights for the attention connection $h$. The softmax-sum module in the illustration corresponds to Eq. \ref{eq:oneshot}, fusing attentions from different blocks. These weights are fully differentiable and are learned together with convolutional filters, enabling the one-shot connectivity search.}
  \label{fig:one-shot}
\end{figure}

For each pair of blocks $(j, i)$ where $L(j) < L(i)$, we place a weight for every $k \in P_{(j, i)}$. Let $h$ be a weight vector of size $m = |P_{(j, i)}|$, and $X^{out}_P = [x_1^{out}, \ldots, x_m^{out}]$ be the tensor concatenating $x_k^{out}$ of every possible peer $k$ in $P$. Then, we reformulate Equation~\ref{eq:attention} as:
\begin{equation}
\label{eq:oneshot}
x_i^{in} = \sum_{(j,i) \in G} \sigma(w_{ji}) \cdot (A(x) \cdot x_j^{out})\quad \mathrm{where}\quad x = \bold{1}^T\left( \mathrm{softmax} \left(h\right) \cdot X^{out}_{P_{(j,i)}} \right).
\end{equation}
$\bold{1}$ is a vector of size $m$ having 1 as all its element values, making $x$ to be a weighted sum of peer block outputs $x_k^{out}$. Use of softmax function allows one-hot like behavior (i.e., selecting one peer to control the attention) based on learned weights $h = [h_1, \ldots, h_m]$. Figure \ref{fig:one-shot} visualizes the process.


The entire process is fully differentiable, allowing the one-shot training to learn the attention weights $h$ together with the connection weights $w_{ji}$. This is unlike AssembleNet which partially relies on exponential mutations to explore connections. Once the attention weights are found, we can either prune the connections by only leaving the argmax over $h_k$ or leave them with softmax. We confirmed that they do not make different in practice, allowing us to only maintain one peer-attention per block as shown in Figure~\ref{fig:peer-attention} (c). Peer-attention only causes 0.151\% increase in computation, which we describe more in Appendix.




\subsection{Model implementation details}
\label{subsec:details}

In order to provide fair comparison to previous work, we comply with the same block structure as AssembleNet \cite{ryoo2019assemblenet}, which by itself is comparable to (2+1)D ResNet-50.

We build two RGB input blocks (whose temporal resolutions are searched), two optical flow input blocks, and one object input block. RGB blocks and optical flow blocks have the same number of channels and layers as AssembleNet, while the object input block only has one max spatial pooling layer which does not increase the number of parameters of the model. 

\begin{figure}
  \centering
  \includegraphics[width=0.49\columnwidth]{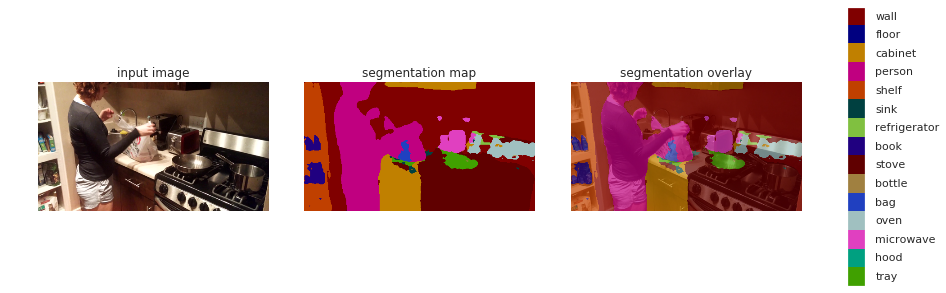}
  \includegraphics[width=0.49\columnwidth]{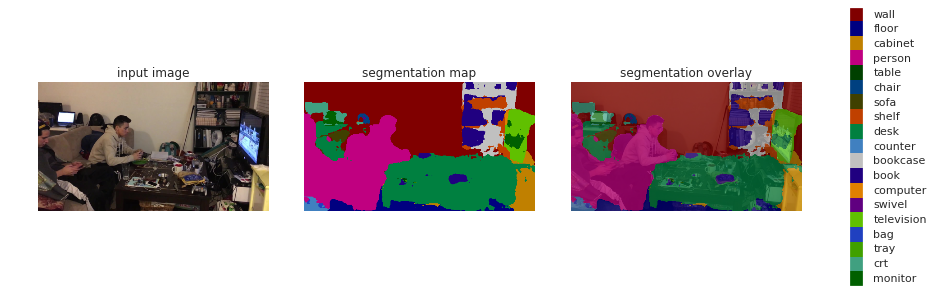}
  \includegraphics[width=0.49\columnwidth]{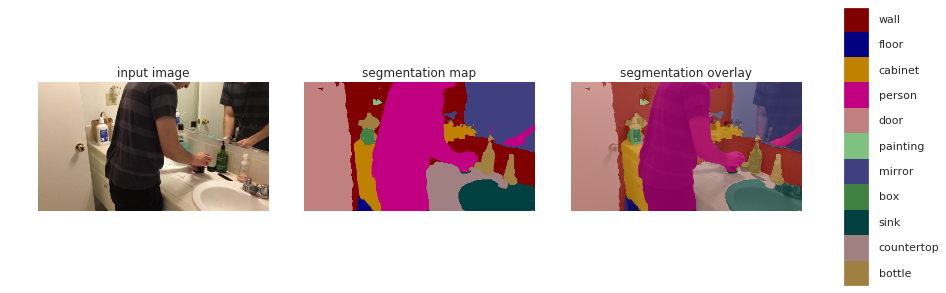}
  \includegraphics[width=0.49\columnwidth]{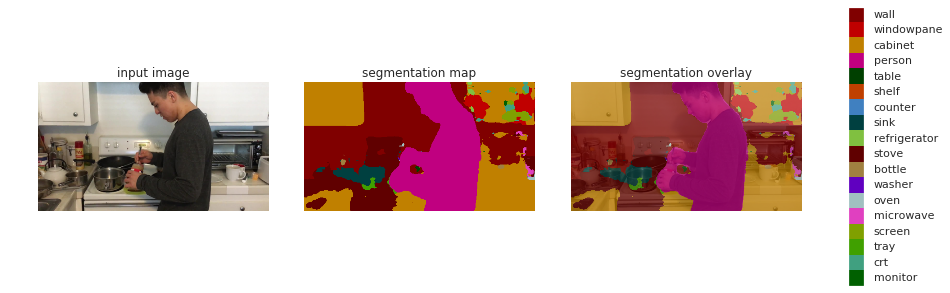}
  \caption{Examples of the segmentation CNN applied directly on Charades video frames with in-home activities. These noisy masks serve as an input to the object input block, suggesting that our video model is required to learn to handle such noisy input.}
  \label{fig:seg-examples}
\end{figure}

The object input block obtains its input from a fixed object segmentation model trained independently with the ADE-20K object segmentation dataset \cite{ade20k}. We treat this module as a blackbox and do not propagate gradients into it.
Because this is an off-the-shelf segmentation module and was not trained on any video dataset, its outputs become noisy when directly applied to video datasets as shown in Figure \ref{fig:seg-examples}.

Our model has convolutional blocks of four levels (five levels if we count input blocks). The sum of channel sizes are held as a constant at each level (regardless the number of blocks), in order to maintain the total number of parameters. The total channels are 128 at input level, and 128, 256, 512, and 512 at levels 1 to 4 following the ResNet module and block formulation. As a result, all models have equivalent number of parameters to standard two-stream CNNs with (2+1)D residual modules.

Each convolutional block was implemented by alternating 2-D residual modules and (2+1)D residual modules as was done in \cite{tran2018closer,ryoo2019assemblenet}. (2+1)D module is composed of 1D temporal convolution layer followed by 2D spatial convolution layer, followed by 1x1 convolution layer. The temporal resolution of each block is controlled using temporally dilated 1-D convolution, avoiding hard frame downsampling. More details of the blocks are in the supplementary material.

Although the number of blocks at each level could be hand-designed, we use AssembleNet architecture search (with an evolutionary algorithm) to find the optimal combination of convolutional blocks and their temporal resolutions. Once we have the blocks, we connect blocks with weighted connections (doing weighted summation) following Section \ref{subsec:connections}. Finally, the one-shot attention search model obtained by implementing our peer-attention with softmax-weighted-sum, as described in Section \ref{subsec:oneshot}.

\subsubsection{Approach summary}
The overall process could be summarized as follows:
\begin{enumerate}
\item Prepare blocks. We use AssembleNet evolution to find convolutional blocks, roughly connected.
\item Initialize our one-shot search model by including all possible block connections as well as new attention connections, as described in Sections \ref{subsec:connections}$\sim$\ref{subsec:oneshot}.
\item Train the one-shot model, learning the attention connectivity weights.
\item Prune low weight connections to make the model more compact. We maintain only one peer-attention per block.
\end{enumerate}

We name our final approach specifically as AssembleNet++.


\section{Experimental results}

We conduct experiments on popular video recognition datasets:
multi-class multi-label Charades~\cite{sigurdsson2018charadesego}, and also the recent Toyota Smarthome dataset~\cite{das2019toyota}, which records natural person activities in their homes. 

We note that we report results
\textbf{without any pre-training} on a large-scale video dataset, which is unlike most of previous work. Regardless of that, AssembleNet++ outperforms prior work. We conduct multiple ablation experiments to confirm the benefit of our multi-modal model formulation with peer-attention and our one-shot attention search.

\subsubsection{Charades dataset.} The Charades dataset~\cite{sigurdsson2018charadesego} is composed of continuous videos of humans interacting with objects. 
This dataset is a multi-class multi-label video dataset with a total of 66,500 annotations. The videos in the dataset involve motion of small objects in real-world home environments, making it a very challenging dataset. Example video frames of the Charades dataset could be found in Figure \ref{fig:seg-examples}. We follow the standard v1 classification setting of the dataset, reporting mAP \%. We use Charades as our main dataset for ablations, as it is a realistic dataset explicitly requiring modeling of interactions between object information and other raw inputs such as RGB.


\subsubsection{Toyota Smarthomes dataset} The Toyota Smarthomes dataset~\cite{das2019toyota} consists of real-world activities of humans in their daily lives, such as reading, watching TV, making coffee or breakfast, etc.  
Humans often interact with objects in this dataset (e.g., `drink from a can' and `cook-cut'). 
The dataset contains 16,115 videos of 31 action classes, and the videos are taken from 7 different camera viewpoints. We only use RGB frames from this dataset, although depth and skeleton inputs are also present in the dataset.


\subsubsection{Baselines}


As a baseline model, we use AssembleNet architecture backbone~\cite{ryoo2019assemblenet} which consists of multiple configurable (2+1)D ResNet blocks. 
Ablations, including our models without the object input block and without peer-attention, are also implemented and compared. 

In our ablation experiments which compare different aspects of the proposed approach 
(Sections \ref{subsec:object-abl}, \ref{subsec:att-abl}, \ref{subsec:exp-abl}, and \ref{subsec:exp-general}), we train the models for 50K iterations with cosine decay for the Charades dataset. When using the Toyota dataset, we train our models for 15K iterations with cosine decay as this is a smaller dataset than Charades (66,500 annotations in Charades vs. 16,115 segmented videos in Toyota Smarthome). Further, when comparing against the state-of-the-art, we use the learning rate following a cosine decay function with `warm-restart' \cite{loshchilov2017}, which we discuss more in Section \ref{subsec:sota}. 

Since the model is a one-shot architecture search to discover the attention connectivity, training is efficient and takes only 20$\sim$30 hours. 


\subsection{Using object modality}
\label{subsec:object-abl}

In this ablation experiment, we explore the importance of the object input. For this study, our model learns the block connectivity from the Charades training, while not using any attention (i.e., they look like Figure \ref{fig:peer-attention} (a)).



\begin{figure}
  \centering
  \includegraphics[width=1.0\columnwidth]{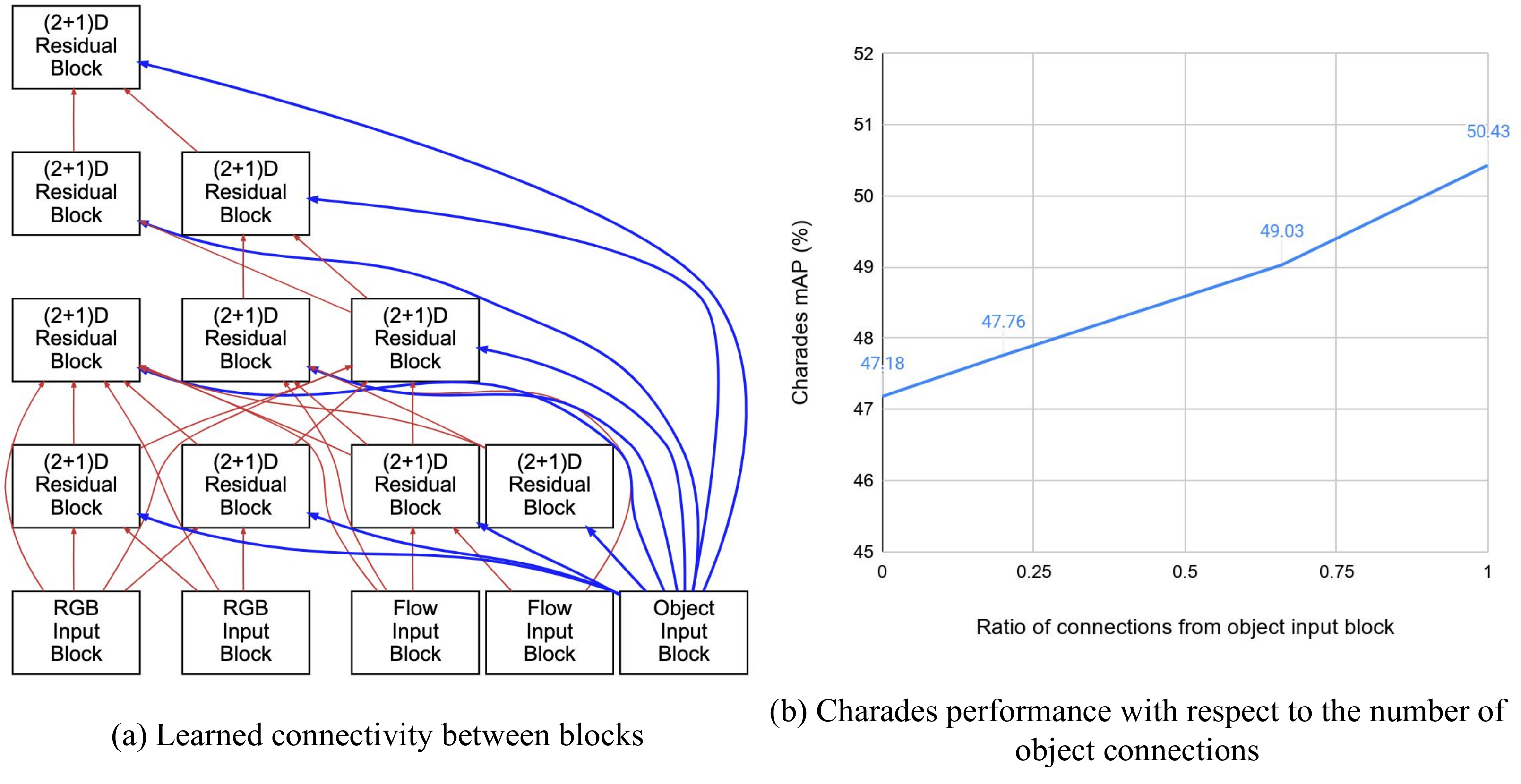}
  \caption{(a) Learned connectivity graph of the model and (b) Charades classification performance per object connection ratio. The highlighted blue edges correspond to the direct connections from the object input block.}
  \label{fig:per-connection-ratio}
\end{figure}

Figure~\ref{fig:per-connection-ratio} (a) shows the best connectivity the one-shot model discovered. This is obtained by (i) evolving the blocks with 100 rounds of architecture evolution, (ii) connecting all blocks, (iii) training the weights in one-shot, and then (iv) pruning the low-weight connections. The connections weights $w_{ji}$ with values higher than 0.2 are visualized. 
Interestingly, the best model is obtained by connecting the object input block to every possible block. The model with this `omnipresent' object connectivity obtains 50.43 mAP on Charades compared to 47.18 mAP of the model without any object connections, which attests the the usefulness of the object modality. The learned weights of each object connection is more than 0.7, suggesting the strong usage of it.

Motivated by the finding that the usage of object information at every block is beneficial (i.e., omnipresent object modality connectivity), we ran an experiment to investigate how performance changes with respect to the best models found with different number of object connections. Figure~\ref{fig:per-connection-ratio} (b) shows the Charades classification performances of our best found models with full vs. restricted object input usage. X-axis of the graph corresponds to how often the model uses the direct input from the object input block. 0 means that it does not use object information at all, and 1 means it fuses the object information at every block. We are able to clearly observe that the performance increases proportionally to the usage of the object information.


\subsection{Attention search}
\label{subsec:att-abl}

Next, we confirm the effectiveness of our proposed AssembleNet++ with attention search. 
Table~\ref{tab:attention} illustrates how much performance gain we get by using attention connections as opposed to the standard weighted connections. 

In addition to attention connectivity with self-attention (Figure \ref{fig:peer-attention} (b)) and peer-attention (Figure \ref{fig:peer-attention} (c)), we implemented and tested `static attention'. This is when learning fixed weights not influenced by any input. We are able to observe that our approach of one-shot attention search (with peer-attention) greatly improves the performance. The benefit was even higher (i.e., by $\sim$6\% mAP) when using the object input.

\begin{table}
\setlength\tabcolsep{5pt}
\caption{Comparison between performance with and without attention connections on Charades (mAP). The models were trained for 50K iterations.}
\label{tab:attention}
\begin{center}
\begin{tabular}{l|c|c}
\hline
Attention &  without object  & ~~~with object~~~ \\ 
\hline
None &  47.18  & 50.43 \\   
Static &  48.82  & 51.15 \\   
Self &  51.91  & 55.40 \\   
Peer &  52.39  & \textbf{56.38} \\   
\hline
\end{tabular}
\end{center}
\end{table}

\subsection{Comparison to the state-of-the-art}
\label{subsec:sota}

In this section, we compare the performance of our AssembleNet++ model with the previous state-of-the-art approaches. We use the model with optimal peer-attention found using our one-shot attention search, and compare it against the results reported by previous work. Unlike most of the existing methods benefiting from pre-training with a large-scale video dataset, we demonstrate that we are able to outperform state-of-the-art \textbf{without such pre-training}. Below, we show our results on Charades and Toyota Smarthome datasets.


We also note that the proposed learned attention mechanisms are very powerful, as also seen in the ablation experiments in Section~\ref{subsec:att-abl}, and even without object information and without pre-training can outperform, or be competitive to, the state-of-the-art.


\subsubsection{Charades dataset}

Table~\ref{tab:charades} shows the results of our method on the Charades dataset. 
Notice that we are establishing a new state-of-the-art number on this dataset, outperforming previous approaches relying on pre-training. Further, we emphasize that our model is organized to have a maximum of 50 convolutional layers as its depth, explicitly denoting it as `AssembleNet++ 50'. Our model performs, without pre-training, even superior to AssembleNet with the depth of 101 layers that uses a significantly larger number of parameters. We also note that the use of object modality and attention mechanism proposed here, improves the corresponding AssembleNet baseline by +12.6\%. 

For this experiment, we use the learning rate with `warm-restart' \cite{loshchilov2017}. More specifically, we use a cosine decay function that restarts to a provided initial learning rate at every cycle. The motivation is to train our models with an identical amount of training iterations compared to the other state-of-the-art. We apply 100K training iterations with two 50K cycles while only using Charades videos, in contrast to previous work (e.g., AssembleNet \cite{ryoo2019assemblenet} and SlowFast \cite{feichtenhofer2018slowfast}) that used 50K pre-training with another dataset and did 50K fine-tuning with Charades on top of it. 
We note that the results of our model without `warm-restart' and without pre-training (as seen in the ablation results in Table~\ref{tab:attention}) at 56.38 are also very competitive to the state-of-the-art. 

\begin{table}[t]
\begin{center}
\caption{Classification performance on the Charades dataset (mAP).}
\label{tab:charades}
\begin{tabular}{|l|l|c|}
\hline
Method &  Pre-training  & mAP \\                         
\hline
Two-stream~\cite{sigurdsson2016asynchronous}  & UCF101 &18.6 \\
CoViAR \cite{wu2018compressed} (Compressed) & ImageNet & 21.9 \\
    Asyn-TF \cite{sigurdsson2016asynchronous} & UCF101 & 22.4 \\
    MultiScale TRN~\cite{zhou2018temporal} (RGB) & ImageNet & 25.2 \\
    I3D \cite{carreira2017quo} (RGB-only) & Kinetics &32.9 \\
    I3D from \cite{wang2018non} (RGB-only) & Kinetics &35.5 \\
    I3D + Non-local \cite{wang2018non} (RGB-only) & Kinetics &37.5 \\
    EvaNet \cite{piergiovanni2018evolving} (RGB-only) & Kinetics & 38.1 \\
    STRG \cite{wang2018videos} (RGB-only) & Kinetics & 39.7 \\
    LFB-101 \cite{wu2018long} (RGB-only) & Kinetics & 42.5 \\
    SGFB-101 \cite{ji2020genome} (RGB-only) & Kinetics & 44.3 \\
    SlowFast-101 \cite{feichtenhofer2018slowfast} (RGB+RGB) & Kinetics & 45.2 \\
    Two-stream (2+1)D ResNet-101  & Kinetics & 50.6 \\
    AssembleNet-50~\cite{ryoo2019assemblenet}  & MiT & 53.0  \\
    AssembleNet-50~\cite{ryoo2019assemblenet}  & Kinetics & 56.6  \\
    AssembleNet-101~\cite{ryoo2019assemblenet}  & Kinetics & 58.6  \\
\hline
AssembleNet-50~\cite{ryoo2019assemblenet} & None & 47.2 \\
\hline
AssembleNet++ 50 (ours) without object~~ & None  & 54.98 \\ 
AssembleNet++ 50 (ours) & None  & \textbf{59.8} \\  
   
\hline
\end{tabular}
\end{center}
\end{table}



\subsubsection{Toyota Smarthome dataset}
\label{sec:results-toyota}

We follow the dataset's Cross-Subject (CS) evaluation setting, and measure performance in terms of two standard metrics of the dataset \cite{das2019toyota}: (1) activity classification accuracy (\%) and (2) `mean per-class accuracies' (\%). Table~\ref{tab:toyota} reports our results. Compared to~\cite{das2019toyota}, which benefits from Kinetics pre-training and additional 3D skeleton joint information, we obtain superior performance while training the model from scratch and without skeletons. We believe we are establishing new state-of-the-art numbers on this dataset.

\begin{table}[t]
\caption{Performance on the Toyota Smarthome dataset. Classification \% and mean per-class accuracy \% are reported. Note that our models are being trained from scratch without any pre-training, while the previous work (e.g., \cite{das2019toyota}) relies on Kinetics pre-training.}
\label{tab:toyota}
\begin{center}
\begin{tabular}{|l|c|c|}
\hline
Method &  Classification \% & ~mean per-class~ \\            
\hline
LSTM \cite{mahasseni2016lstm} & - & 42.5 \\
I3D (with Kinetics pre-training) & 72.0 & 53.4 \\
I3D (pre-trained) + NL \cite{wang2018non} & - & 53.6 \\
I3D (pre-trained) + separable STA \cite{das2019toyota} & 75.3 & 54.2 \\
\hline
Baseline AssembleNet-50  & 77.77 & 57.42  \\
Baseline + self-attention & 77.59 & 57.84 \\
\hline
Ours (object + self-attention) & 79.08 & 62.30 \\
Ours (object + peer-attention) & \textbf{80.64} &\textbf{63.64} \\
\hline
\end{tabular}
\end{center}
\end{table}

\subsection{Ablation}
\label{subsec:exp-abl}

\begin{table}
\setlength\tabcolsep{5pt}
\caption{Comparing AssembleNet++ using peer-attention vs. a modification using 1x1 convolutional layer instead of attention. They use an identical number of parameters. Charades classification accuracy (mAP) and Toyota mean per-class accuracy (\%) are reported.}
\label{tab:ablation}
\begin{center}
\begin{tabular}{l|c|c}
\hline
Model &  Charades & ~~Toyota~~ \\            
\hline
Base & 50.43 & 59.16 \\
Base + 1x1 conv. & 50.24 & 59.44 \\
Random peer-attention & 53.40 & 60.23 \\
Our peer-attention & \textbf{56.38} & \textbf{63.64} \\
\hline
\end{tabular}
\end{center}
\label{tab:exp-abl}
\end{table}


In this experiment, we explicitly compare AssembleNet++ using peer-attention with its modifications using the same number of parameters. Specifically, we compare our model against (i) the model using 1x1 convolutional layers instead of attention and (ii) the model using peer-attention but with random attention connectivity. For (i), we make the number of 1x1 convolutional layer parameters identical to the number of parameters in FC layers for attention. Table \ref{tab:exp-abl} compares the accuracies of these models on Charades and Toyota Smarthome datasets. While using the identical number of parameters, our one-shot peer-attention search model obtains superior results.


\subsection{General applicability of the findings}
\label{subsec:exp-general}

Based on the findings that (1) having `omnipresent' neural connectivity from the object modality and (2) using attention connectivity are beneficial, we investigate further whether such findings are generally applicable for many different CNN models. We add object modality connections and attention to (i) standard R(2+1)D network, (ii) two-stream R(2+1)D network, (iii) original AssembleNet, and (iv) our Charades-searched network (without object connectivity and attention), and observe how their recognition accuracy changes compared to the original models. Our model without object and attention is obtained by manually removing connections from the object input block.

Table~\ref{table:generality} shows the results tested on Charades. We are able to confirm that our findings are applicable to other manually designed, as well as, architecture searched architectures. The increase in accuracy is significant for all architectures. Note that our architecture itself is not significantly superior to AssembleNet without object.  However, since its connectivity was searched together the object input block (i.e., Section \ref{subsec:oneshot}), we are able to observe that our model better takes advantage of the object input via peer attention.
50K training iterations with cosine decay was used for this comparison.


\begin{table}
\setlength\tabcolsep{5pt}
\caption{Comparison between original CNN models (without object modality and without attention) and their modifications based on our attention connectivity and object modality. The value corresponding to `AssembleNet++' for the column `base' is obtained by manually removing connections from the object input block and removing attention from our final one-shot attention search model. Measured with Charades classification (mAP, higher is better), trained from scratch for 50k iterations.}
\label{table:generality}
\begin{center}
\begin{tabular}{l|c|c}
\hline
CNN model~ &  ~~base~~  & + object~+ attention \\ 
\hline
RGB R(2+1)D &  36.51  & \textbf{45.30} \\  
Two-stream R(2+1)D &  39.93  & \textbf{47.74} \\   
AssembleNet &  47.18  & \textbf{53.48} \\  
AssembleNet++  &  47.62  & \textbf{56.38} \\
\hline
\end{tabular}
\label{tab:generality}
\end{center}
\end{table}

\section{Conclusion}

We present a family of novel video models which are designed to learn interactions between the object modality input and the other raw inputs: AssembleNet++. We propose connectivity search to fuse new object input into the model, and introduce the concept of peer-attention to best capture the interplay between different modality representations. The concept of peer-attention generalizes previous channel-wise self-attention by allowing the attention weights to be computed based on other intermediate representations. An efficient differentiable one-shot attention search model is proposed to optimize the attention connectivity.
Experimental results confirm that (i) our approach is able to appropriately take advantage of the object modality input (by learning connectivity to the object modality consistently) and that (ii) our searched peer-attention greatly benefits the final recognition. The method outperforms all existing approaches on two very challenging video datasets with daily human activities.
Furthermore, we confirm that our proposed approach and the strategy are not just specific to one particular model but is generally applicable for different video CNN models, improving their performance notably.




\par\vfill\par

\clearpage
%
%
\bibliographystyle{splncs04}
\bibliography{bib}
\end{document}


\pagestyle{headings}
\mainmatter
\def\ECCVSubNumber{3586}  

\title{AssembleNet++: Assembling Modality Representations via Attention Connections \\- Supplementary Material -}

\titlerunning{AssembleNet++}
%
\author{Michael S. Ryoo\inst{1,2} \and
AJ Piergiovanni\inst{1} \and
Juhana Kangaspunta\inst{1} \and \\Anelia Angelova\inst{1}}
\authorrunning{M. S. Ryoo et al.}
%
\institute{Robotics at Google \and
Stony Brook University\\
\email{\{mryoo,ajpiergi,juhana,anelia\}@google.com}}
\maketitle

\appendix
\section{Appendix}

\subsection{Convolutional blocks with (2+1)D residual modules}

Each convolutional block is implemented by alternating 2-D residual modules and (2+1)D residual modules as was done in \cite{ryoo2019assemblenet}. Each (2+1)D residual module does 1D temporal convolution first, and then 2D spatial convolution followed by 1x1x1 convolution. This (2+1)D residual module is also similar the ones used in \cite{feichtenhofer2018slowfast}.
We use the filter size of 3x3 for spatial convolutional layers, and the size of 3 for temporal convolutional layers.
Temporal dilation from \cite{ryoo2019assemblenet} was used to control the temporal resolution of each (2+1)D block.
2D and (2+1)D residual modules are repeated multiple times in each block. As a result, our residual blocks have a total of 9, 12, 18, and 9 convolutional layers for levels 1 to 4, making the depth of the network comparable to ResNet-50.

Since we follow the AssembleNet (2+1)D block design, the total number of filers in each level is maintained as a constant, regardless the number of blocks in the level. 
That is, our model has the number of parameters equivalent to the two-stream version of ResNet-50.

Each RGB input block has 1 spatial convolutional layer (filter size 7x7, stride 2x2), 1 temporal convolutional layer (filter size 5, stride 1), and one max pooling layer (pool and stride size 2x2). Each optical flow input block has 1 spatial convolutional layer (filter size 7x7, stride 2x2) and one max pooling layer (pool and stride size 2x2). The object input block has only one max pooling layer (pool and stride size 4x4).

\subsection{Computation overhead of peer-attention}

Our approach is adding very little computation overhead. As was done in previous differentiable architecture search \cite{bender2018understanding,liu2019darts}, once the one-shot search is finished and the attention connection weights (i.e., $h$) are obtained, only a single peer node is selected by the softmax for each block and the others are discarded. Figure 1(c) in the paper shows an example of the final model.

As a result, similar to what was reported in \cite{hu2018squeeze} with channel-wise self-attention, our peer-attention only causes 0.151\% increase in the total computation (FLOPs). The increase in the number of parameters is 1.68\%. Peer-attention only adds one fully connected layer after every block. Each FC layer has an identical number of parameters as a 1x1 conv layer, while spending significantly less amount of computation (compared to 1x1) as it does not have spatial resolution.

\subsection{Learned architecture}

We also provide our model in table form in Table~\ref{tab:model}. In particular, the 3rd column of the table shows the connectivity: a list of blocks where the input to that block is coming from. The fusion of the convolutional block outputs with different tensor shapes is done using a spatial pooling and a 1x1 conv layer, before the weighted summation. Further, notice that (as described in Section 3.5) there is the one-shot peer-attention search module implemented before every input to a convolutional block (i.e., Figure 2 in the paper), in addition to Table \ref{tab:model}. The final block (i.e., block 14) is followed by a FC layer to generate logits. Temporal max pooling is used to combine logits of different frames in videos. Finally, cross entropy loss is used to train the network.

\begin{table}
    \centering
    \setlength\tabcolsep{5pt}
    \caption{The table form of our model with detailed parameters. This model corresponds to Figure 4 in the paper. ``$C$, dilation, stride'' in the Table correspond to the ResNet channel parameter, temporal dilation rate, and spatial stride. Note that the object input block does not have any convolutional layer, and 151 is the number of object categories which decide the size of its input channel.}
    \label{tab:model}
    \begin{tabular}{|l|l|c|p{26mm}|}
    \hline
      Index & Level & Input connections & Block parameters: \newline $C$, dilation, stride\\
    \hline
    0 & 0 & [RGB] & 32, 2, 4 \\ \hline
    1 & 0 & [RGB] & 32, 4, 4 \\ \hline
    2 & 0 & [Flow] & 32, 1, 4 \\ \hline
    3 & 0 & [Flow] & 32, 1, 4 \\ \hline
    4 & 0 & [Object] & 151, 1, 4 \\ \hline
    5 & 1 & [0, 1, 2, 3, 4] & 32, 1, 1 \\ \hline
    6 & 1 & [0, 1, 4] & 32, 4, 1 \\ \hline
    7 & 1 & [2, 3, 4] & 32, 8, 1 \\ \hline
    8 & 1 & [2, 3, 4] & 32, 1, 1 \\ \hline
    9 & 2 & [0, 1, 2, 4, 5, 6, 7, 8] & 64, 4, 2 \\ \hline
    10 & 2 & [2, 3, 4, 7, 8] & 64, 1, 2 \\ \hline 
    11 & 2 & [0, 4, 5, 6, 7] & 128, 8, 2 \\ \hline
    12 & 3 & [4, 11] & 256, 8, 2 \\ \hline
    13 & 3 & [2, 3, 4, 5, 6, 7, 8, 10, 11] & 256, 4, 2 \\ \hline
    14 & 4 & [4, 12, 13] & 512, 2, 2 \\
    \hline
    \end{tabular}
\end{table}

\par\vfill\par

\clearpage
%
%
\bibliographystyle{splncs04}
\bibliography{bib}